\documentclass[]{llncs}
\usepackage[T1]{fontenc}
\usepackage{graphicx}
\usepackage{hyperref}

\usepackage{amsmath}
\usepackage{gensymb}
\usepackage{subcaption}
\usepackage{colortbl}
\usepackage{xcolor}

\usepackage{array,booktabs}
\newcolumntype{L}[1]{>{\raggedright\let\newline\\\arraybackslash\hspace{0pt}}m{#1}}
\newcolumntype{C}[1]{>{\centering\let\newline\\\arraybackslash\hspace{0pt}}m{#1}}
\newcolumntype{R}[1]{>{\raggedleft\let\newline\\\arraybackslash\hspace{0pt}}m{#1}}

\begin{document}
\title{Assessing Robustness to Catastrophic Forgetting in Continual Deep Reinforcement Learning for Cyber-Physical Systems}
%
%
\author{Kim N. Nolle\inst{1} \and
Ivana Dusparic\inst{1} \and
Rhodri Cusack\inst{2} \and
Vinny Cahill\inst{1}}
\authorrunning{F. Author et al.}
%
\institute{School of Computer Science and Statistics, Trinity College Dublin, Dublin, Ireland
\and Trinity College Institute of Neuroscience, Trinity College Dublin, Dublin, Ireland
}
\maketitle              
%

\begin{abstract} 
As the use of reinforcement learning (RL) becomes more wide\-spread in cyber-physical systems, it becomes increasingly important to design RL agents that are able to adapt to changes in their environments in order to maintain safety and reliability. This paper addresses the question of whether different classes of RL algorithm are more robust against forgetting previously learned behaviours, a critical issue in continual learning, in non-stationary environments. We evaluate the robustness of representative examples of three important classes of RL algorithm in a non-stationary autonomous parking environment. These are Proximal Policy Optimization (PPO), which is a \textit{model-free on-policy} algorithm, Soft-Actor-Critic (SAC), which is a \textit{model-free off-policy} algorithm, and Drama, which is a \textit{model-based} algorithm. The agents must learn to successfully park in four different scenarios, which are presented successively, with parking spaces oriented at varying angles. We find off-policy algorithms appear to be more robust compared to on-policy algorithms due to their use of replay, while model-based algorithms do not appear to offer any direct advantage over model-free algorithms. These experiments also highlight the additional challenges of finding robust abstractions of the environment, and oversensitivity to hyperparameters.

\end{abstract}
\textcolor{red}{This preprint has not undergone peer review or any post-submission improvements or corrections. The Version of Record is published in \textit{Artificial Intelligence XLIII} and is available online at \url{https://doi.org/10.1007/978-3-032-45236-8\_34}.}

\section{Introduction}
Reinforcement learning (RL) has made incredible progress in recent years and is becoming increasingly prevalent in cyber-physical systems, such as autonomous robotics, vehicles, and drones  \cite{Govinda_2025_RL-AutonomousSystems}. These are very dynamic domains where agents may encounter diverse environments potentially requiring new skills. Even something as seemingly straightforward as autonomous parking can exhibit significant diversity (e.g. parking spaces can have different angles and widths, and there may be oncoming traffic or pedestrians that do not always behave predictably). However, RL agents are often trained on a narrow set of tasks and can struggle to adapt to new tasks and/or changes in the target environment. Even small changes in the environment can cause RL agents to suffer significant drops in performance. This can potentially compromise the safety and reliability of these systems as they are deployed to the real world.

Continual reinforcement learning (CRL) aims to address RL in non-stationary environments \cite{Khetarpal_2022}. In contrast to conventional machine learning, which typically learns from stationary data, continual learning learns from non-stationary data distributions \cite{Wang_2024_CLreview}. 
This means that, as the data distribution changes, agents should be able to incrementally gain new abilities while retaining and potentially updating previously learned abilities. 

A well-known problem in CRL is catastrophic forgetting, which is a phenomenon where previously acquired abilities are lost as new abilities are learned \cite{Khetarpal_2022}. 
Existing CRL research focuses on proposing mechanisms to prevent catastrophic forgetting, which are typically built on top of existing RL algorithms (see \cite{Khetarpal_2022} for a review). However, there is little discussion about the suitability of the underlying RL algorithms themselves for non-stationary environments. This research expands on the work in \cite{Nolle_2025_CRL} and investigates whether different classes of RL algorithm are more robust to forgetting and thus offer a more suitable basis for designing algorithms for non-stationary environments.

To this end, we evaluate the robustness of three different classes of RL algorithm in a non-stationary autonomous parking environment. In particular, we evaluate representative examples of three important classes of RL algorithm: Proximal Policy Optimization (PPO) \cite{Schulman_2017_PPO}, which is a \textit{model-free on-policy} algorithm, Soft-Actor-Critic (SAC) \cite{Haarnoja_2018_SAC}, which is a \textit{model-free off-policy} algorithm, and Drama \cite{Wang_2025_Drama}, which is a \textit{model-based} algorithm. In the autonomous parking environment, the agents must learn to park in four different scenarios with parking spaces oriented at varying angles. The agent is successively trained in these four scenarios one after another, representing a non-stationary environment.
The results suggest that that off-policy algorithms are more robust to forgetting compared to on-policy algorithms because their use of replay allows data from past tasks to be reused during optimization. 
In addition, the results show that an interleaved-training regime improves the agents' ability to learn all the scenarios, although this presupposes knowledge about future task distributions.
These experiments also expose additional challenges that are not commonly discussed in the CRL literature, such as finding abstractions of the environment that generalize to diverse environments, and the oversensitivity to hyperparameters.

\section{Background and Related Work}

RL is a machine learning paradigm in which a so-called `agent' learns a behaviour by learning how to maximize a numerical reward as a result of interacting with its environment \cite{SuttonBarto_2018}. These interactions can be formalized as a Markov Decision Process (MDP), which is defined by a state space $\mathcal{S}$, an action space $\mathcal{A}$, a set of rewards $\mathcal{R}$ and a transition probability function $p$. The behaviour of an agent is defined by a policy $\pi(a|s)$, which is a probability distribution over possible actions $a\in\mathcal{A}$ given the current state $s\in\mathcal{S}$.

A crucial distinguishing characteristic of RL algorithms is whether or not they use a model of the environment, also known as a world model \cite{SuttonBarto_2018,Hafner_2025_DreamerV3}.
\textit{Model-free} methods do not rely on world models and iteratively update their policy based on experience gathered through direct interaction with the environment. \textit{Model-based} methods train a world model to simulate experience, which can be used for planning or to train a model-free RL algorithm. World models essentially learn the transition function and reward function of the environment. In principle, the same world model may support a range of different tasks \cite{Hafner_2025_DreamerV3}, suggesting that model-based approaches are a good basis for continual learning.

One way of classifying model-free RL algorithms is based on how training data is generated. During the training of an RL agent, there is a distinction between the \textit{behaviour policy} and the \textit{target policy} \cite{Wang_2024_DRLsurvey}. The behaviour policy is the policy that the agent follows when generating training data. The target policy is the policy that is optimized through learning and should maximize the returns. \textit{On-policy} methods use the target policy as the behaviour policy. This means that in each round of optimization, the target policy is used to generate the training samples. At the end of each round, these samples are discarded to avoid optimising old policies.
In \textit{off-policy} methods on the other hand, the target and behaviour policies differ. The samples are generated by the behaviour policies and are stored in a buffer. Samples from this buffer are then used to update the target policy. On-policy algorithms have low variance and fast convergence properties, while off-policy methods have better data efficiency \cite{SuttonBarto_2018,Wang_2024_DRLsurvey}.  A priori, the convergence properties of on-policy algorithm might suggest that they might offer a good basis for continual learning.

Continual reinforcement learning (CRL) is defined as RL in non-stationary environments \cite{Khetarpal_2022}. This means that the components of the MDP that underlie the environment can change over time. This can include gradual changes in the environment or sudden changes. 
A special case of this is piecewise non-stationarity. Here, the non-stationary function $f(i,t)$ can be broken up into time intervals where the function is stationary, such as:
\begin{equation}
    \label{equ:piecewise-non-stationary}
    f(i, t) = \left\{\begin{array}{lr}
        f_0(i) & \text{for } 0\leq t < t_0\\
        f_1(i) & \text{for } t_0\leq t < t_1\\
        f_2(i) & \text{for } t_1\leq t < t_2\\
        ...
        \end{array}\right.
\end{equation}
An example of this is a multi-task setting, where each task $z$ is defined by a stationary MDP $\mathcal{M}^{(z)} = \langle \mathcal{S}^{(z)}, \mathcal{A}^{(z)}, \mathcal{R}^{(z)}(s, a), p^{(z)}(s, a) \rangle$. Piecewise non-stationarity then describes the scenario in which an agent encounters tasks one after another.

CRL algorithms, and CL algorithms in general, are often categorized into \textit{architecture-based}, \textit{regularization-based}, and \textit{rehearsal-based} algorithms \cite{Khetarpal_2022,Wang_2024_CLreview}. Ar\-chi\-tecture-based methods dynamically modify the architecture of the neural network to accommodate new knowledge. One approach to doing this is to increase the size of the network as new tasks are learned \cite{Rusu_2016_PNN,Malagon_2024_CompoNet}. Regularization-based methods attempt to prevent forgetting by constraining updates to representations. This can be done, for example, by adding regularization terms to the loss functions used for neural network training to balance performance in old tasks with that in new tasks \cite{Wang_2024_CLreview}. Rehearsal-based methods store previous experiences that are presented to the agent again during training. While the buffer in off-policy algorithms is typically implemented as a simple FIFO queue, rehearsal-based algorithms propose mechanisms to create better representations of the true long-term data distribution in the buffer \cite{Khetarpal_2022}. Despite the wide range of proposed algorithms, CRL, especially for real-world applications, remains an open challenge and an active area of research \cite{Khetarpal_2022,Wang_2024_CLreview,Nolle_2025_CRL}.

\section{Experimental Design}

\begin{figure}[h]
    \centering
    \begin{subfigure}{0.25\textwidth}
        \centering
        \includegraphics[width=\textwidth]{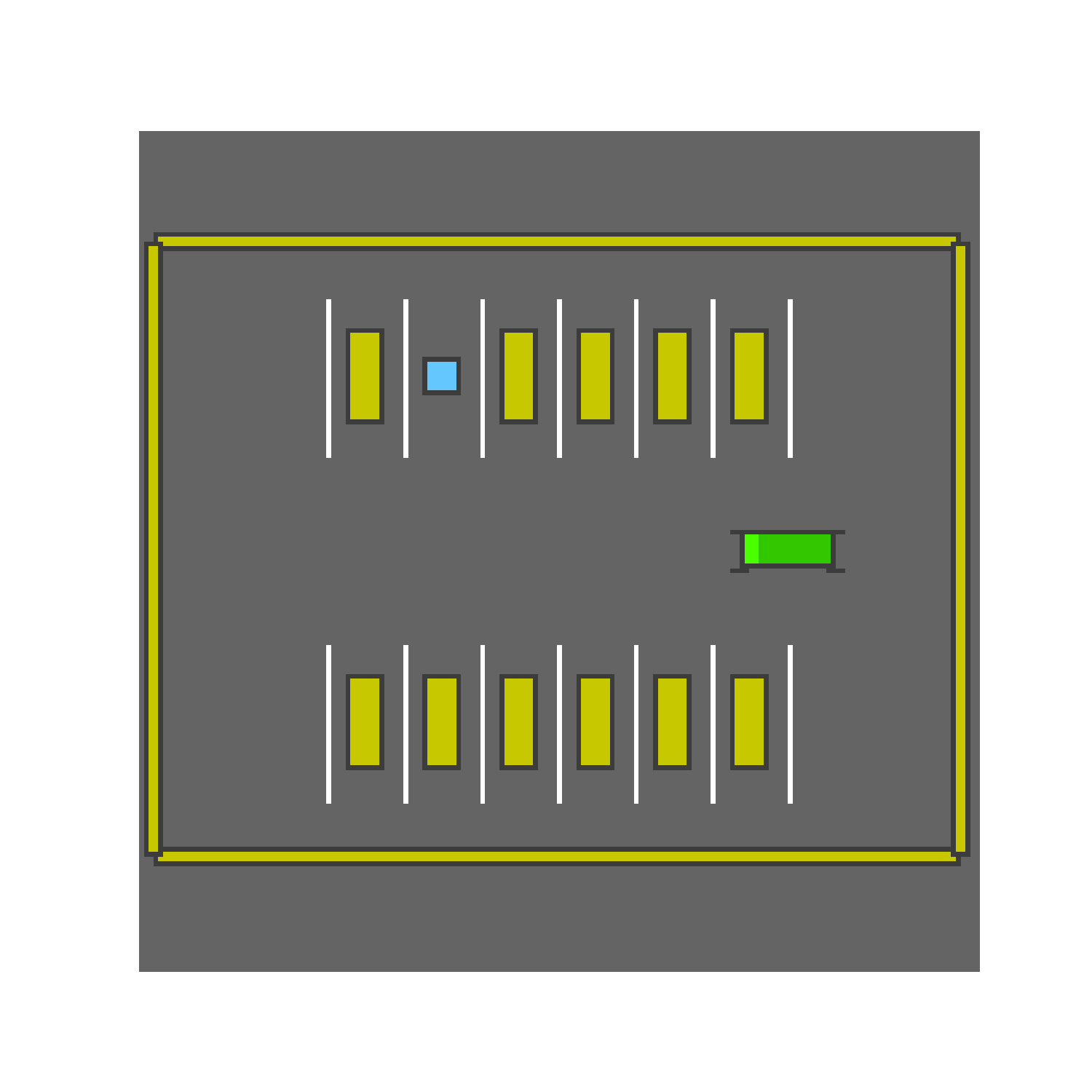}
        \caption{Perpendicular}
    \end{subfigure}\hfill\begin{subfigure}{0.25\textwidth}
        \centering
        \includegraphics[width=\textwidth]{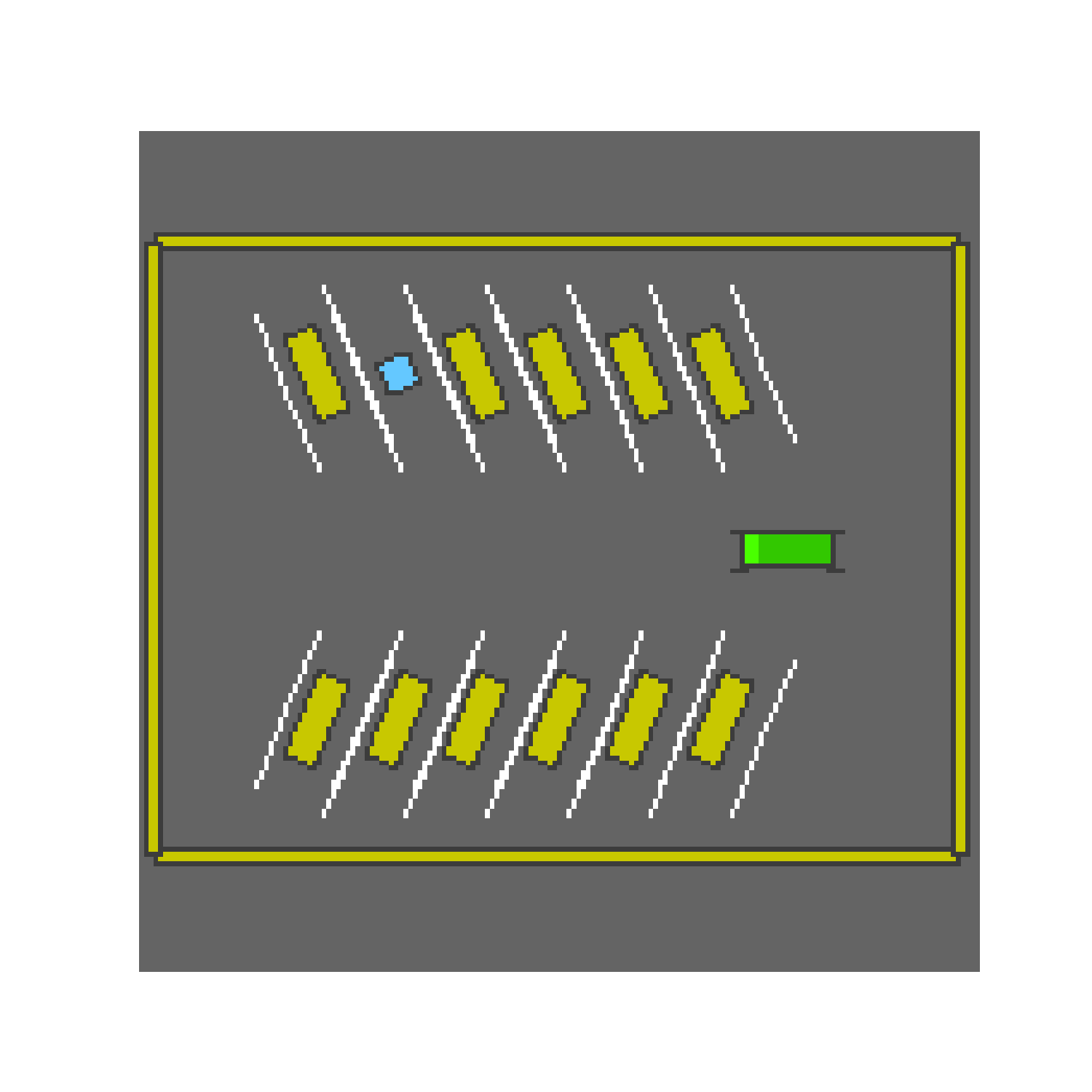}
        \caption{25\degree diagonal}
    \end{subfigure}\hfill\begin{subfigure}{0.25\textwidth}
        \centering
        \includegraphics[width=\textwidth]{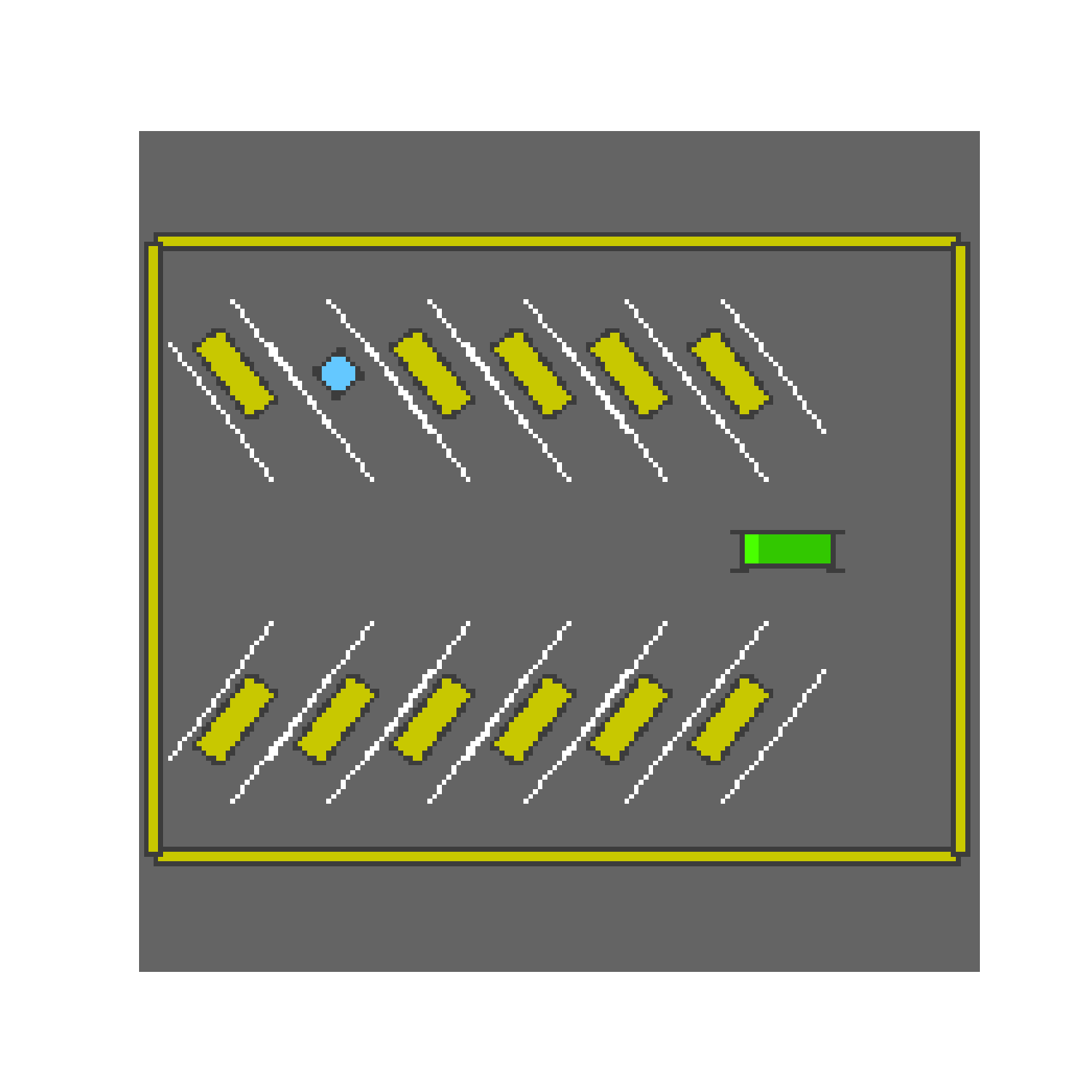}
        \caption{50\degree diagonal}
    \end{subfigure}\hfill\begin{subfigure}{0.25\textwidth}
        \centering
        \includegraphics[width=\textwidth]{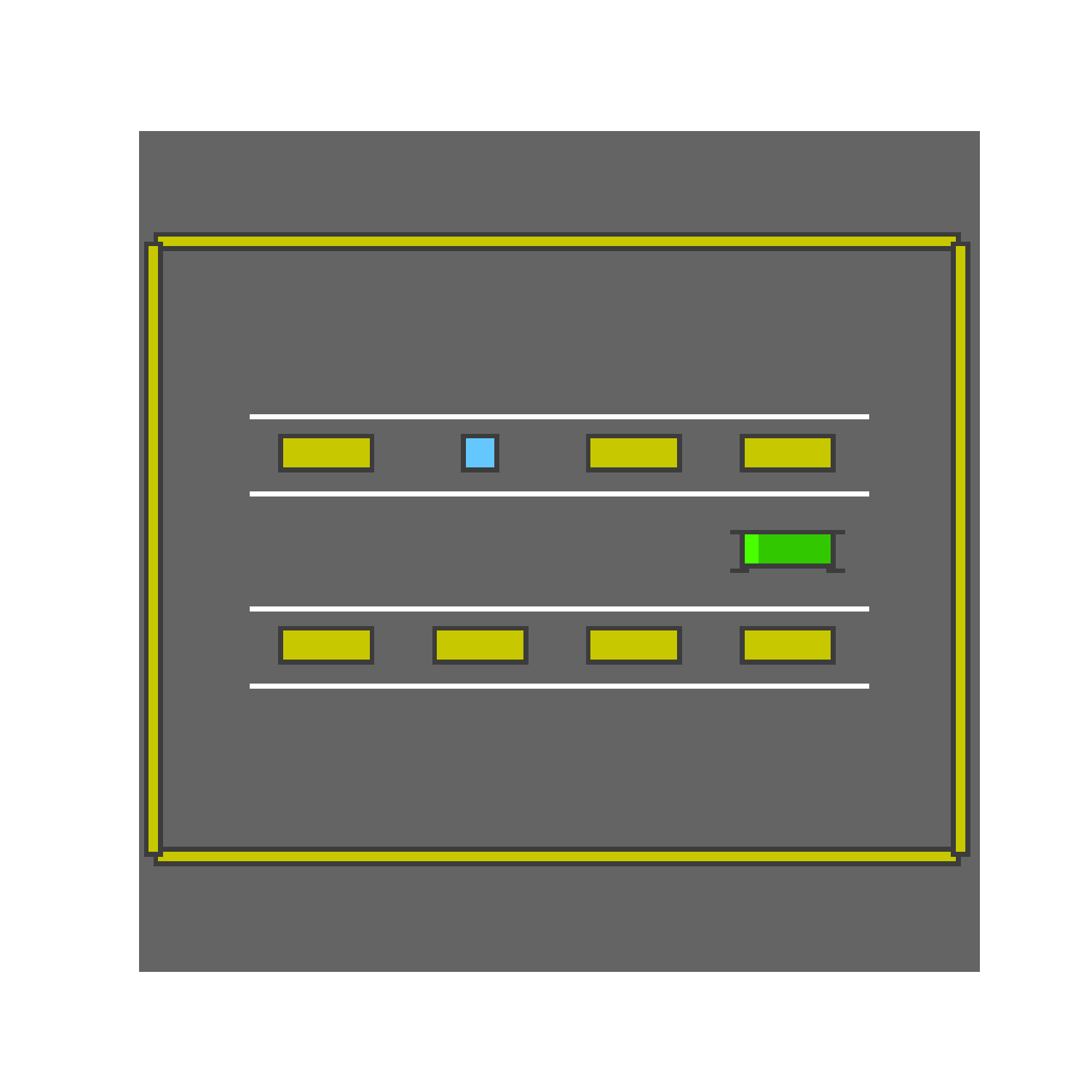}
        \caption{Parallel}
    \end{subfigure}
    \caption{The four different parking tasks. The green rectangle represents the agent vehicle and the blue square represents the goal parking space.}
    \label{fig:modified_env}
\end{figure}

An autonomous driving environment based on the \texttt{highway-env} python package \cite{highway-env} was used to explore CRL. The parking environment is an episodic goal-based environment in which the agent should park a vehicle in a specified parking space (the goal). There are three possible outcomes of an episode: the agent successfully parks in the goal parking space, the agent crashes, or the agent fails to complete the task within the maximum allowed number of time steps. Based on this we defined four tasks: (i) perpendicular parking, (ii) 25\degree~diagonal parking, (iii) 50\degree~diagonal parking, and (iv) parallel parking. These different tasks vary only in the angle of the parking spaces (see Fig. \ref{fig:modified_env}). Fundamentally, these tasks should be very similar and so it might be expected that an RL agent would be able to learn to park in all four scenarios, one after the other and without forgetting, readily. This represents a piecewise non-stationary environment where the state space changes over time. 

As explained, we consider three state-of-the-art algorithms that are representative of different classes of RL algorithm. PPO (model-free, on-policy) uses a clipped objective function, which constrains potentially destabilising updates resulting in more stable training \cite{Schulman_2017_PPO}. SAC (model-free, off-policy) provides sample-efficient, stable learning. Drama (model-based) uses Mamba \cite{Gu_2024_Mamba} as the sequence-modelling backbone for the world model, which, given our computational constraints, enables use of a significantly smaller model that achieves comparable performance to other state-of-the-art model-based algorithms \cite{Wang_2025_Drama}. Here, Drama trains a PPO agent in the world model. Our code is publicly available at \url{https://github.com/knolle1/cl-parking}.

The hyperparameters of the RL agents were tuned in the perpendicular parking task using random search and the same hyperparameters were used in all four scenarios. Reward engineering was also conducted for the perpendicular parking task and this reward model was used for all four tasks. This setup foresees a scenario in which future tasks are not known beforehand and so are not considered during hyperparameter tuning and reward engineering.

Various metrics have been proposed for assessing CRL. In addition to reporting the learning curves to show how the average performance of the agent in each task evolves over time, we also consider the metrics for backward transfer (BWT) and forward transfer (FWT) proposed by \cite{Lopez-Paz_2017_GEM}: 
\begin{equation}
    \label{equ:BWT}
    BWT=\frac{1}{T-1}\sum^{T-1}_{i=1}R_{T,i}-R_{i,i}
\end{equation}
\begin{equation}
    \label{equ:FWT}
    FWT=\frac{1}{T-1}\sum_{i=2}^TR_{i-1,i}-\overline{b}_i
\end{equation}
$T$ is the total number of tasks, $R_{i,j}$ is the performance in task $i$ after learning task $j$, and $\overline{b}_i$ is the  performance of a randomly initialized model in task $i$.

\section{Results}

As a baseline, we first train the tasks individually. 
We then evaluate sequential training of the four tasks in the order (i) - (iv) and the effect of reversing the task order. We also investigate training tasks in an interleaved manner, using Elastic Weight Consolidation (EWC) \cite{Kirkpatrick_2017_EWC} to prevent forgetting, and the challenge of applying a reward model to a non-stationary environment.

\subsection{Baselines}
\label{sec:baselines}

\begin{figure}[!h]
    \centering
    \begin{subfigure}{0.7\textwidth}
        \centering
        \includegraphics[width=\textwidth]{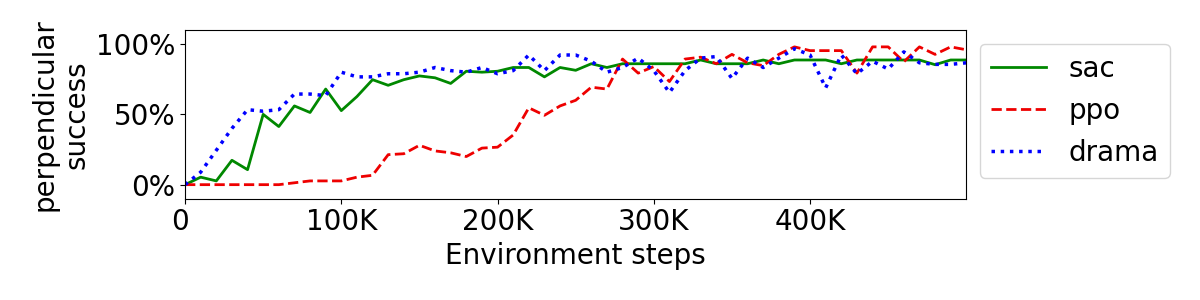}
        \caption{Perpendicular parking}\label{fig:single-training-success:perpendicular}
    \end{subfigure}\\
    \begin{subfigure}{0.7\textwidth}
        \centering
        \includegraphics[width=\textwidth]{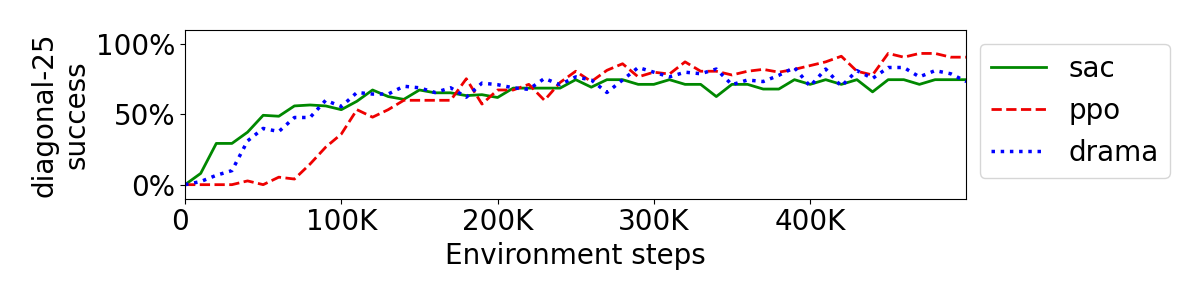}
        \caption{25\degree diagonal parking}\label{fig:single-training-success:25diagonal}
    \end{subfigure}\\
    \begin{subfigure}{0.7\textwidth}
        \centering
        \includegraphics[width=\textwidth]{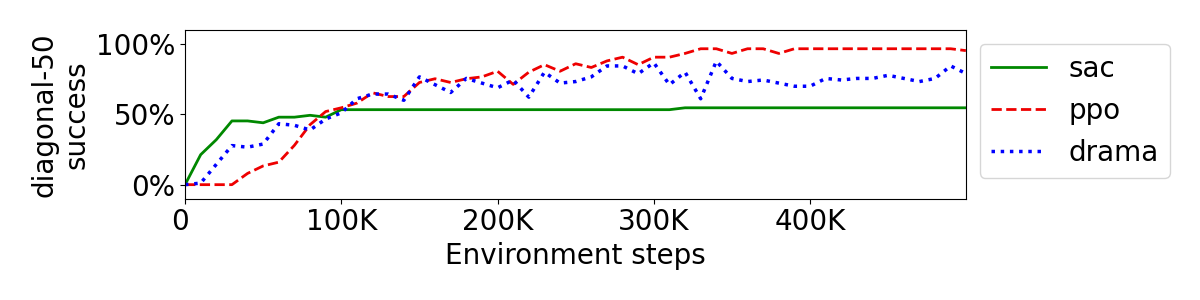}
        \caption{50\degree diagonal parking}\label{fig:single-training-success:50diagonal}
    \end{subfigure}\\
    \begin{subfigure}{0.7\textwidth}
        \centering
        \includegraphics[width=\textwidth]{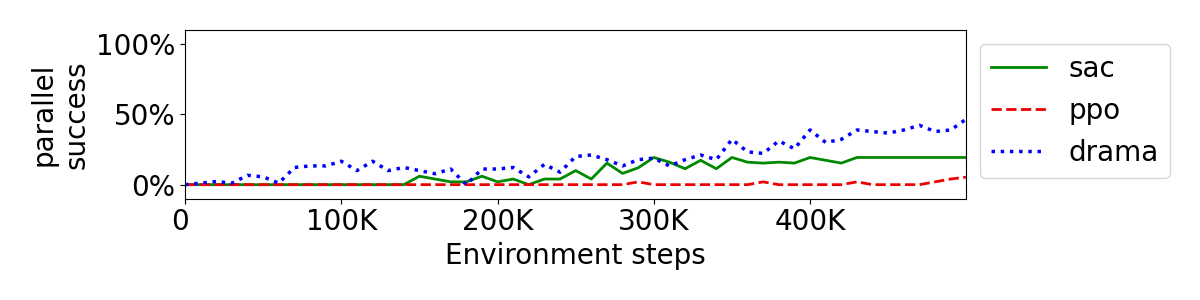}
        \caption{Parallel parking}\label{fig:single-training-success:parallel}
    \end{subfigure}
    \caption{Comparison of episode success throughout single task training of the four parking tasks (averaged over 5 runs for PPO and SAC; 3 runs for Drama)}
    \label{fig:single-training-success}
\end{figure}

We initially train the PPO, SAC, and Drama agents on each task individually. Every 10,000 steps throughout training, the current policy of the agent is evaluated on 30 episodes of the task and the success rate is calculated. Fig. \ref{fig:single-training-success:perpendicular} shows the resulting learning curves. It can be seen that SAC and Drama converge faster than PPO, which may be due to the increased sample efficiency of these algorithms. The off-policy nature of SAC means that past experiences from the replay buffer can be reused for calculating update gradients, as opposed to on-policy algorithms like PPO, which require new samples to be collected for each gradient step \cite{Haarnoja_2018_SAC}. Model-based RL algorithms like Drama improve sample efficiency by generating artificial training samples \cite{Wang_2025_Drama}.

\begin{figure}[h]
    \centering
    \begin{subfigure}{0.25\textwidth}
        \centering
        \includegraphics[width=\textwidth]{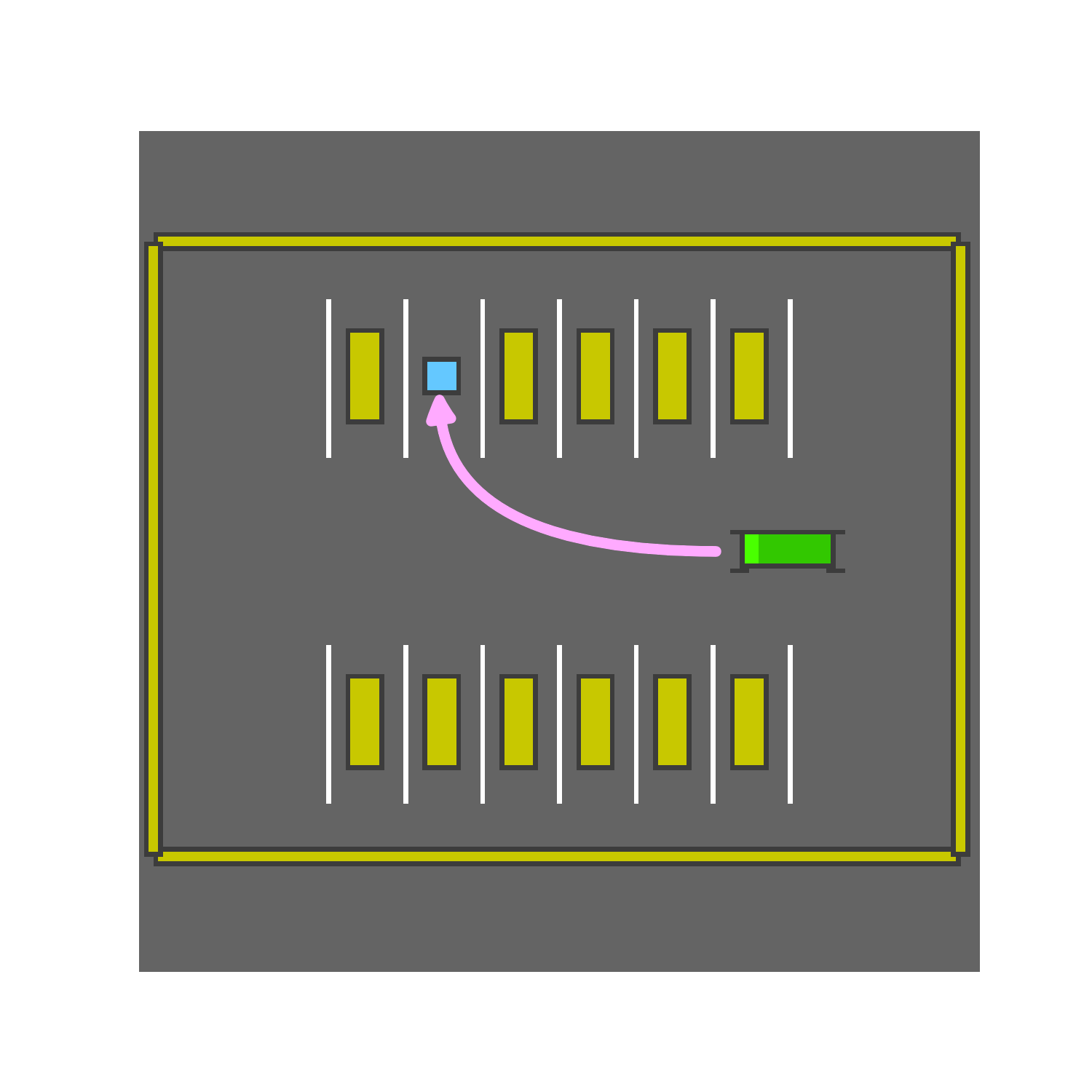}
        \caption{Perpendicular}
    \end{subfigure}\hfill\begin{subfigure}{0.25\textwidth}
        \centering
        \includegraphics[width=\textwidth]{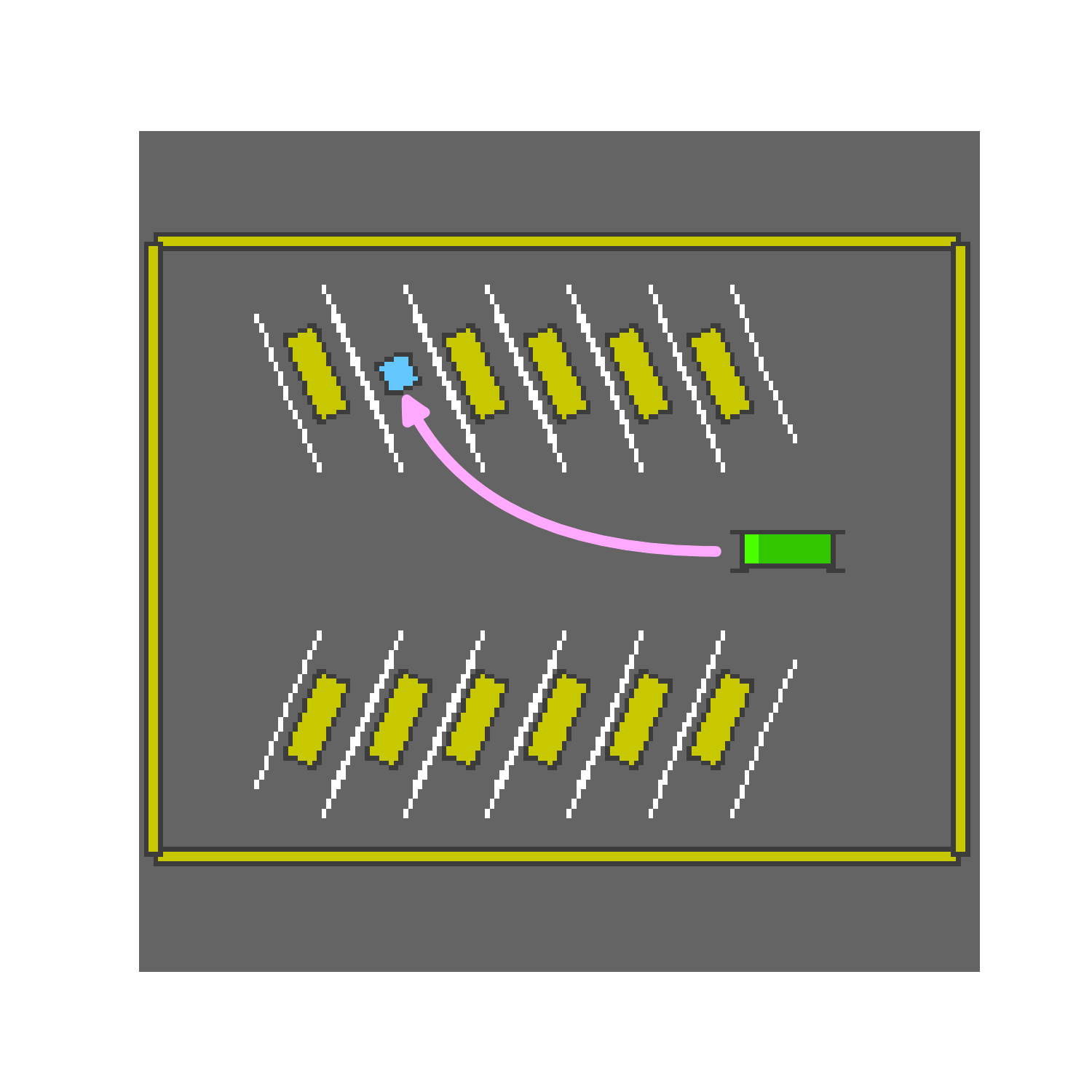}
        \caption{25\degree diagonal}
    \end{subfigure}\hfill\begin{subfigure}{0.25\textwidth}
        \centering
        \includegraphics[width=\textwidth]{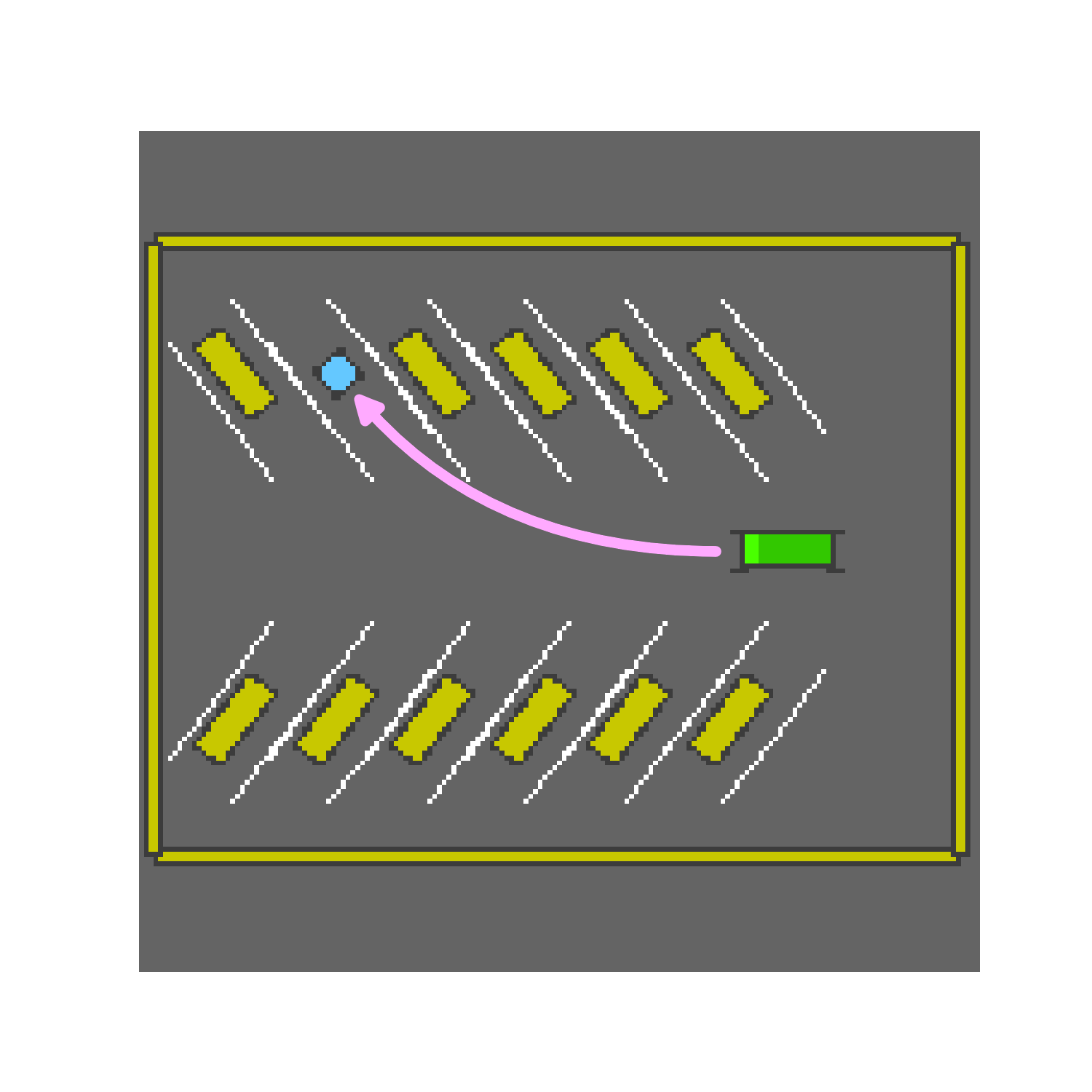}
        \caption{50\degree diagonal}
    \end{subfigure}\hfill\begin{subfigure}{0.25\textwidth}
        \centering
        \includegraphics[width=\textwidth]{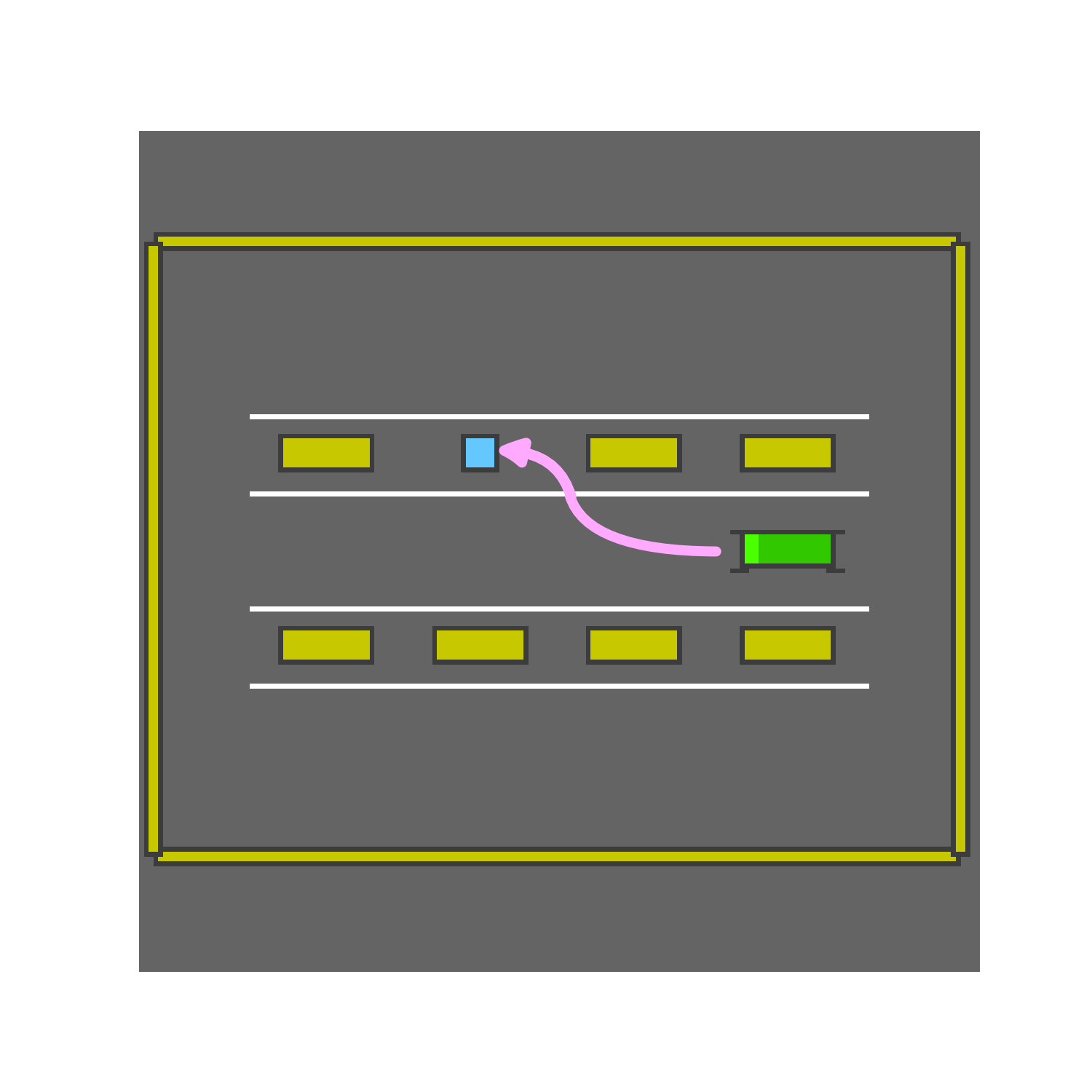}
        \caption{Parallel}
    \end{subfigure}
    \caption{Expected trajectories of the controlled car in the different parking tasks}
    \label{fig:expected-trajectory}
\end{figure}

Notably, all three RL algorithms struggle to learn parallel parking (Fig. \ref{fig:single-training-success:parallel}). A possible reason is that parallel parking is, in fact, less similar to the other tasks than originally anticipated. Fig. \ref{fig:expected-trajectory} shows the expected trajectories for the different tasks, where it becomes clear that perpendicular and diagonal parking are expected to have similar trajectories while parallel parking is expected to require more manoeuvring.

\subsection{Sequential Task Training}
\label{sec:results:sequential}

\begin{figure}[!h]
    \centering
    \begin{subfigure}{0.5\textwidth}
        \centering
        \includegraphics[width=\textwidth]{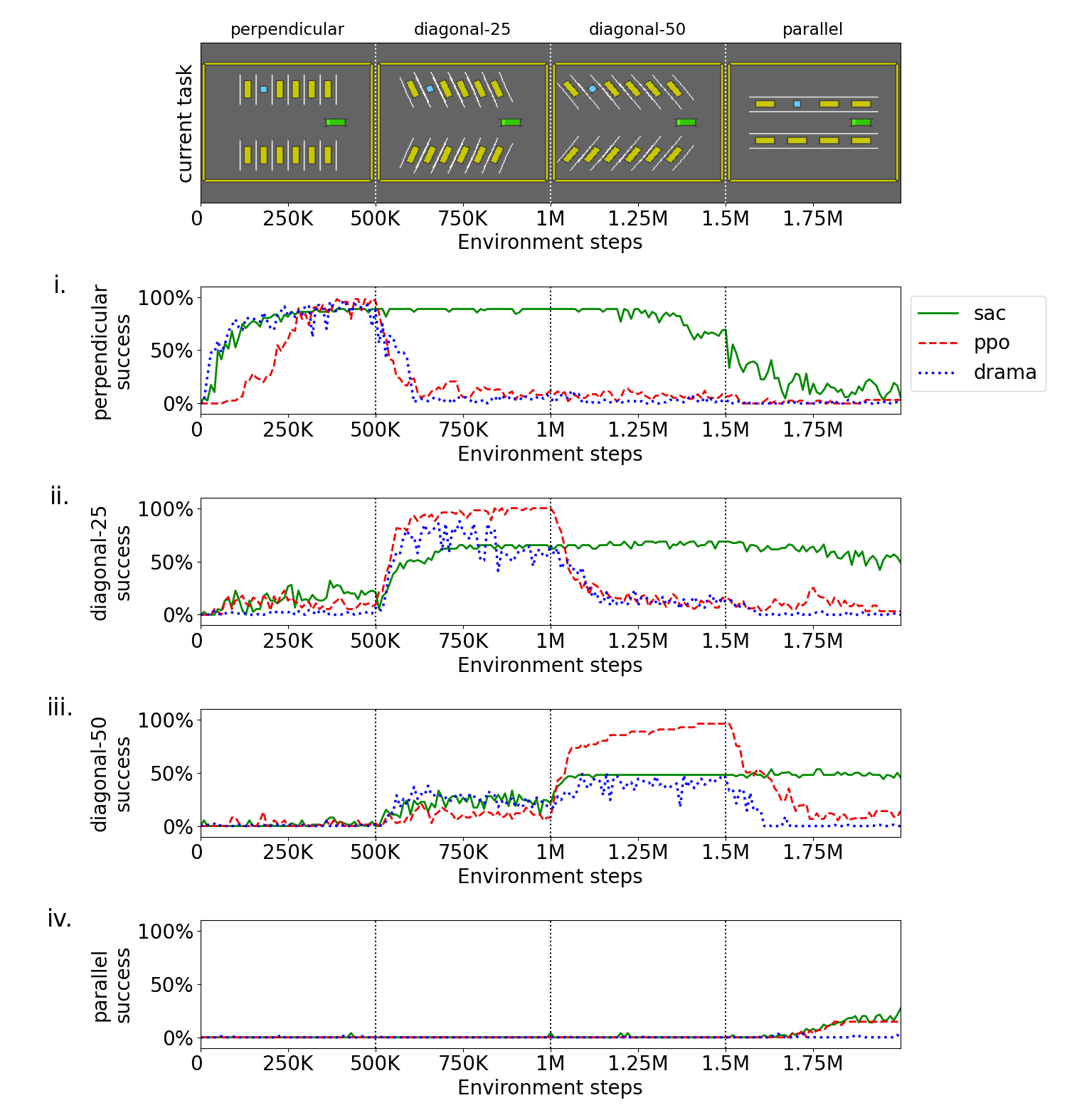}
        \caption{Initial task sequence}\label{fig:sequential-training-success:inc}
    \end{subfigure}\hfill\begin{subfigure}{0.5\textwidth}
        \centering
        \includegraphics[width=\textwidth]{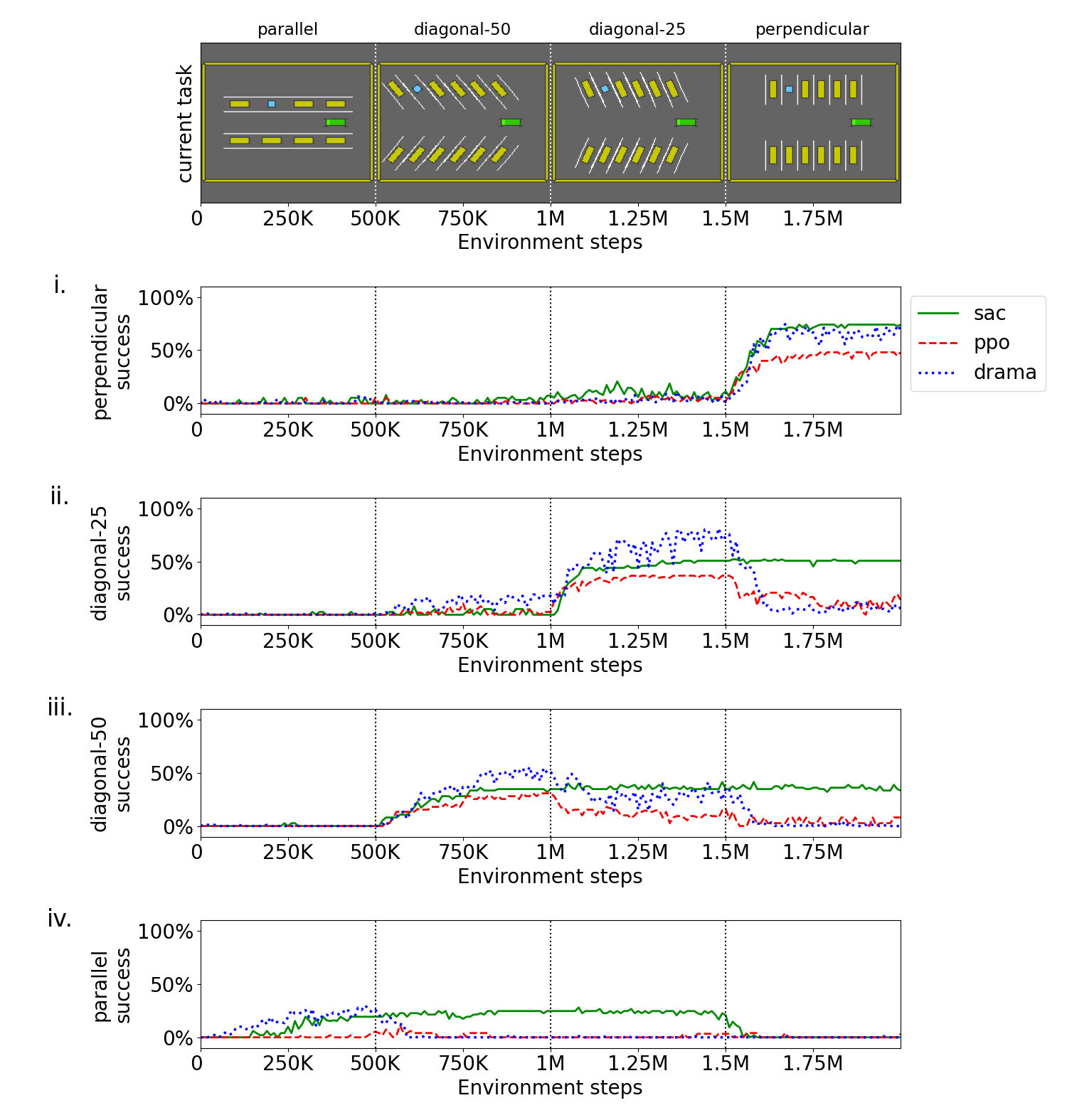}
        \caption{Reverse task sequence}\label{fig:sequential-training-success:dec}
    \end{subfigure}
    \caption{Comparison of episode success in each task throughout sequential task training (averaged over 5 runs for PPO and SAC; 3 runs for Drama)}
    \label{fig:single-training-success}
\end{figure}

Next, we train the RL algorithms on all four tasks in sequence. Each scenario is trained for 500k steps. The success rate of the agents in all four tasks is evaluated throughout training and can be seen in Fig. \ref{fig:sequential-training-success:inc}.
In the first 500k steps, the agents are being trained on the perpendicular parking task. Fig. \ref{fig:sequential-training-success:inc}(i) shows that the performance of the agents in perpendicular parking during this phase behaves the same as in the single-task experiments (Fig. \ref{fig:single-training-success:perpendicular}), which is to be expected. Fig. \ref{fig:sequential-training-success:inc}(ii) also shows that there is a slight improvement in 25\degree~diagonal parking, even though the agents have at this point in time not been exposed to this task yet. This suggests some positive forward transfer between perpendicular and 25\degree~diagonal parking.

After the first task switch from perpendicular parking to 25\degree~diagonal parking at 500k steps, the performance of PPO and Drama in perpendicular parking (Fig. \ref{fig:sequential-training-success:inc}(i)) drastically decrease, while their performance in the new task (25\degree~diagonal parking) increases (Fig. \ref{fig:sequential-training-success:inc}(ii)). This clearly demonstrates catastrophic forgetting, as previously acquired abilities are lost as new abilities are learned \cite{Khetarpal_2022}. 
In contrast, SAC is able to maintain performance in perpendicular parking for longer and only starts to lose performance around step 1.25 million (Fig. \ref{fig:sequential-training-success:inc}(i)).
An explanation for this might be the replay buffer, which has a size of 1 million elements. This means that up to step 1 million, the data from perpendicular parking is still stored in the replay buffer. The agent is effectively being jointly optimized on perpendicular parking and 25\degree diagonal parking, and thus can maintain performance on perpendicular parking. After this, the buffer is full and the oldest samples are replaced by new data. As a result, there is increasingly less data from perpendicular parking used to optimize the agent and so the agent forgets this task.

Next, we investigate the effect of reversing the task order. Fig. \ref{fig:sequential-training-success:dec} plots the the success rates of the learned policy in order of decreasing parking angle (parallel, 50\degree~diagonal, 25\degree~diagonal, and perpendicular parking). Similar to the results in \ref{fig:sequential-training-success:inc}, PPO and Drama show signs of forgetting after task switches while SAC is able to maintain performance as long as sufficient samples of past tasks remain in the replay buffer. However, a clear difference is that the agents achieve lower success rates in the diagonal and perpendicular parking tasks compared to the results in single task training and the sequential training starting with perpendicular parking. One interpretation is that attempting to learn the parallel parking task induces a bias that makes it more difficult to learn later tasks.

\subsection{Interleaved Task Training}

We also evaluate the effect of interleaving the tasks. At the end of each training episode, the task for the next episode is sampled from a uniform distribution across all four tasks. This represents a conventional multi-task setting where the different tasks are essentially trained in parallel. Here, the agents are trained for a total of 2 million steps so that, on average, each task is visited for 500k steps. The converged success rates should therefore be comparable with the success rates in the single task training experiments.

\begin{figure}[h]
    \centering
    \includegraphics[width=0.8\linewidth]{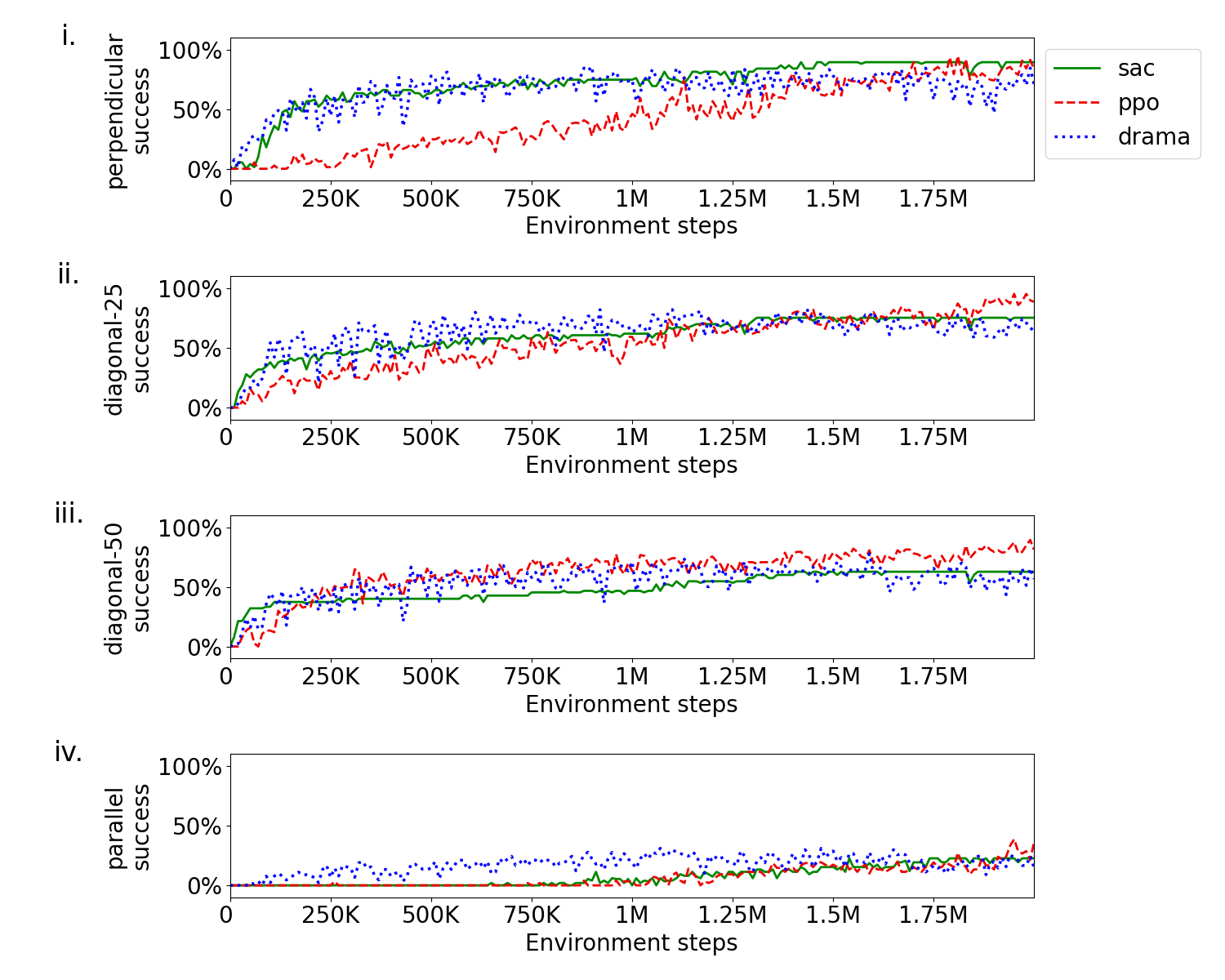}
    \caption{Comparison of episode success in each task throughout interleaved task training (averaged over 5 runs for PPO and SAC; 3 runs for Drama)}
    \label{fig:interleaved-training-success}
\end{figure}

Fig. \ref{fig:interleaved-training-success} shows the results for interleaved training. The plots show that the RL algorithms converge to success rates similar to those in the single task experiments (Fig. \ref{fig:single-training-success}). This suggests that the agents all have the capacity to represent policies for all four tasks simultaneously that are comparable to the agents trained in the single task setting. Furthermore, this lends credibility to the idea that a replay buffer that is large enough to contain a sufficient amount of data for all visited tasks can retain all learned policies.

\subsection{Using EWC to Prevent Forgetting}
\label{sec:results:ewc}

The sequential training experiments clearly demonstrated catastrophic forgetting in all three classes of agent. The results for SAC show the benefit of a replay buffer to preventing forgetting. While the replay buffer retains a sufficient amount of samples from past tasks, the policy for these tasks can be retained. This highlights the potential of replay-based approaches of CRL.
However, the use of such CRL algorithms may not be appropriate for on-policy algorithms like PPO as they store past experiences and replay them \cite{Khetarpal_2022}. On-policy algorithms generate the training samples using the target policy \cite{Wang_2024_DRLsurvey}. This means that the experiences stored in the replay buffer would have been generated by outdated policies rather than the current policy. As a result, the policy gradients resulting from the experiences in the replay buffer might not be indicative of how to optimize the current policy. This is reflected in the fact that CRL literature that proposes rehearsal-based algorithms tends to use off-policy algorithms 
\cite{Daniels_2022_GR,Riemer_2019_SRM,Atkinson_2021_RePR}.

Instead, we explore using Elastic Weight Consolidation (EWC) \cite{Kirkpatrick_2017_EWC}, to prevent catastrophic forgetting in the PPO agent. 
EWC is a well-known regularization based CRL algorithm that is often used as a baseline in CRL literature, e.g.,  \cite{Riemer_2019_SRM,Atkinson_2021_RePR,Schwarz_2017_ProgCompr,Traore_2019_DisCoRL}. EWC constrains updates to parameters that are considered to be important to previously learned tasks \cite{Kirkpatrick_2017_EWC}. For each previously learned task, a quadratic penalty proportional to the importance of the parameters is calculated and these penalties are added to the loss of the task currently being learned. The importance of parameters is estimated using the Fisher information matrix. 
The resulting loss can be seen in (\ref{equ:ewc_loss}), where $\mathcal{L}_T(\theta)$ is the loss of the current task $T$, $F^t$ is the Fisher information matrix for a previous task $t$, $\theta_i$ are the current parameters, and $\theta^*_{t,i}$ are the learned optimal parameters for the previous task $t$. The hyperparameter $\lambda$ weighs how important the past tasks are compared to the current task.
\begin{equation}
    \label{equ:ewc_loss}
    \mathcal{L}(\theta)=\mathcal{L}_T(\theta)+ \sum^{T-1}_{t} \sum_i \frac{\lambda}{2}F^t_i(\theta_i-\theta^*_{t,i})^2
\end{equation}
A limitation of this approach is that the Fisher information matrix and the optimal parameters have to be stored for each task, which means that the memory usage increases with each new task. This is addressed by an online variant \cite{Schwarz_2017_ProgCompr}, where the regularization terms are aggregated:
\begin{equation}
    \label{equ:online-ewc_loss}
    \mathcal{L}(\theta)=\mathcal{L}_T(\theta)+\sum_i \frac{\lambda}{2}F^{total}_i(\theta_i-\theta^*_{i})^2
\end{equation}
\begin{equation}
    \label{equ:ewc-update}
    F^{total}_T \gets \gamma_{EWC}\ F^{total}_{T-1} + F_{T}
\end{equation}
Only the aggregated Fisher information matrix $F^{total}$ and the most recent learned optimal parameters $\theta^*$ need to be stored, resulting in constant memory requirements. The hyperparameter $\gamma_{EWC}$ can be used to decay past regularizers in a controlled manner. 

\begin{table}[!h]
    \centering
    \caption{Forward and backward transfer metrics for sequential training of PPO with and without EWC}
    \label{tab:transfer-metrics-success:ppo-ewc}
    \vspace{5pt}
    \begin{tabular}{C{1.5cm}C{1.5cm}C{1.5cm}C{1.5cm}}
    \toprule
        $\mathbf{\lambda}$  & $\mathbf{\gamma_{EWC}}$ & \textbf{BWT} & \textbf{FWT}\\
    \midrule
    \multicolumn{2}{c}{Without EWC} &  -0.904 & 0.058 \\
    \arrayrulecolor{gray}\hline
       0.5 & 0.9 & -0.818 & 0.067 \\
       1   & 0.9 & -0.813 & 0.062 \\
       10  & 0.9 & -0.820 & 0.031 \\
       \hline
       1 & 0.1 & -0.773 & 0.062 \\
       1 & 0.5 & -0.764 & 0.062 \\
       1 & 0.9 & -0.813 & 0.062 \\
    \arrayrulecolor{black}\bottomrule
    \end{tabular}
\end{table}

Table \ref{tab:transfer-metrics-success:ppo-ewc} shows the BWT and FWT metrics of the PPO agent with EWC in sequential training (initial task sequence) using different EWC hyperparameters. These results show that, while the use of EWC increases the BWT compared to vanilla PPO, which means that there is less forgetting, the BWT is still negative. This means that EWC is unable to fully solve the problem of catastrophic forgetting. 
It can also be seen that varying these hyperparameters did not significantly improve the effect of EWC on catastrophic forgetting. As the $\lambda$ increases, FWT decreases. This may be because increasing $\lambda$ increases the strength of regularization, which constrains updates to the network more, thus making learning new tasks more difficult. Other authors \cite{Traore_2019_DisCoRL} have similarly struggled to find hyperparameters for EWC that prevents forgetting without regularising the network so strongly that the agent is unable to learn new tasks.

\subsection{Solving Parallel Parking}

\begin{figure}[!h]
    \centering
    \includegraphics[width=0.7\linewidth]{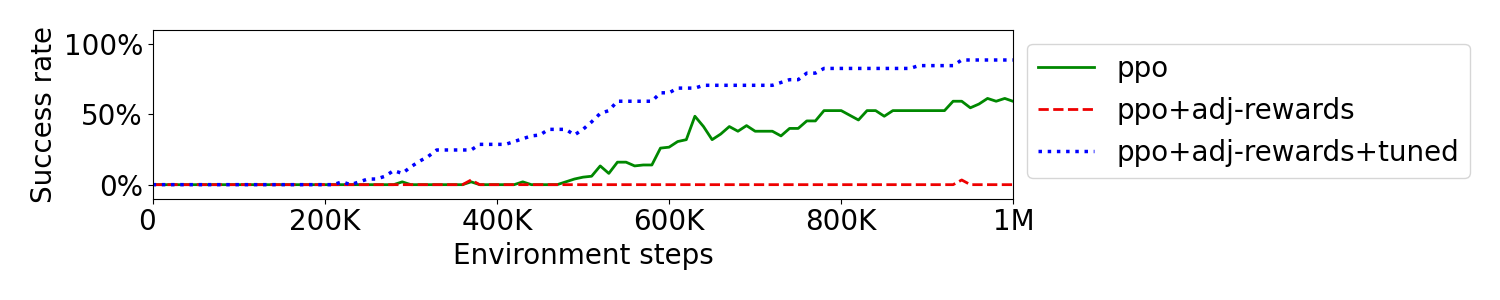}
    \caption{Episode success in additional parallel parking experiments}
    \label{fig:parallel-training-success:1M}
\end{figure}

As discussed in section \ref{sec:baselines}, all three types of RL agent struggle to learn parallel parking. To better understand why this is, additional experiments with PPO were conducted. At first it was hypothesized that longer training time or conducting hyperparameter tuning on the parallel parking task might alleviate the problem. This however was not the case. Fig. \ref{fig:parallel-training-success:1M} (solid line) shows that even with longer training (1 million steps), the success rate converges to around 60\%. Furthermore, we were unable to find hyperparameters that resulted in achieving 100\% success rate. Instead modifications to the reward model had to be made for the agent to successfully parallel park.

This reward model in previous experiments was designed to facilitate learning in the perpendicular parking task. It is proportional to the difference between the agent's state and the goal state, and the agent incurs a penalty when it collides with obstacles. The ability of agents to learn diagonal parking using this reward model shows that this reward model transfers well to the diagonal parking tasks.
However, the agent struggles to learn parallel parking with this reward model.

The reason for this is that the heading angle of the vehicle is included in the state representation. When the difference between the agent and goal angle is small, then the reward is high and vice versa. 
However, this means when the agent is pointing in the opposite direction to the goal in parallel parking, the agent would have to turn 180\degree~to maximize the reward, even though functionally, this situation should be the same as when the goal is pointing in the same direction as the agent. The agent has to make a 180\degree~turn, making the task more difficult.

To account for this, the reward model was modified so that being rotated 180\degree~from the goal produces the same reward as when the agent is facing the same direction. The results of this can be seen in Fig. \ref{fig:parallel-training-success:1M} (dashed line). While the success rate remains low, recordings of the episodes showed that the agent learns to move close to the goal but stops next to the parking space. Additional hyperparameter tuning with this reward model results in an agent that successfully learns the parallel parking task, which can be seen in Fig. \ref{fig:parallel-training-success:1M} (dotted line). This included different learning rates, larger experience buffers, and larger minibatch sizes.

\section{Discussion}

Applying RL to non-stationary environments is made difficult by a number of challenging problems, which were demonstrated by our experiments. The results clearly show catastrophic forgetting in all three types of RL agent. Reversing the task order highlighted the impact of biases introduced by early tasks can have on future learning and the resulting sensitivity to task order. These challenges are related to the so-called \textit{stability-plasticity dilemma}. The stability-plasticity dilemma refers to the trade-off between learning from recent experiences (plasticity) and retaining knowledge from past experiences (stability) \cite{Khetarpal_2022}. Catastrophic forgetting suggests that agents are too plastic and require more stability. On the other hand, plasticity is required in order to overcome biases from past tasks and to learn new tasks. 

These experiments also exposed other challenges that are not commonly discussed in CRL literature.
The first is the oversensitivity of RL, and neural network training in general, to hyperparameters. While the initial hyperparameters worked well for perpendicular and diagonal parking, they did not lead to successful parallel parking. 
It is therefore important to consider how to select hyperparameters that are robust to all future tasks, particularly if not all future tasks are known beforehand.
Another challenge is defining abstractions of the environment that are robust to non-stationarity. The parallel parking experiments showed that, while the reward model designed for solving perpendicular parking transferred well to the diagonal parking tasks, it was not suitable for successfully learning parallel parking. 
This may apply to other abstractions of the environment, such as the state space or action space. 
All three evaluated RL algorithms suffered from this dependence on reward models and hyperparameters.  

The interleaved training experiments showed that this training regime was more successful at learning all four parking tasks compared to the sequential training settings. However, if applied to real world applications, this approach requires the future task distributions to be known at training time.

Off-policy algorithms appear to have some benefits over on-policy algorithms. In particular, off-policy algorithms can reuse old training data. Not only does this improve sample efficiency, but it also means that the agent can continue to be trained on data from past tasks. As shown in section \ref{sec:results:sequential}, this can delay and slow down forgetting. The interleaved training results support the idea that an agent with a replay buffer large enough to contain sufficient data from all visited tasks can retain all learned policies. 
Furthermore, there does not appear to be any obvious benefit of using model-based approaches except for improving sample efficiency due to training in imagination. A possible reason is that, while the world model learns the dynamics of the environment, which may be generalizable across all tasks, the learned policy is still task-specific, which makes it susceptible to the same catastrophic forgetting observed in the model-free variant (PPO).

Overall, the results show that RL algorithms on their own are not suited for non-stationary environments and require additional mechanisms to overcome problems like catastrophic forgetting. Architecture-based CRL algorithms are able to fully prevent forgetting by freezing network parameters for past tasks and increasing the size of the network to accommodate learning new tasks \cite{Rusu_2016_PNN,Malagon_2024_CompoNet}. However, this comes at cost of resource efficiency, with the memory requirements growing as new tasks are learned. Regularization-based algorithms like EWC are able to improve resource efficiency by maintaining a constant network size and balance the stability-plasticity dilemma by adding regularization terms to the loss function. However, as seen in section \ref{sec:results:ewc}, finding suitable hyperparameters for balancing the task loss and regularization terms can be difficult. 
The results for the SAC agent and the interleaved training show the potential of using replay to preventing forgetting. 
However, the main limitation of rehearsal-based methods is the capacity of the replay buffer; replay is only useful if it retains a sufficient amount of samples from past tasks. This poses a problem in real-world applications where the number of tasks may be large but memory is limited. Some existing works have attempted to compress the information in the replay buffer into a generative model, such as a  generative adversarial network (GAN), which can be sampled to generate data for replay \cite{Atkinson_2021_RePR}. However, these approaches only appear to work well for small numbers of tasks, with the capacity of the generative model again being the limiting factor \cite{Atkinson_2021_RePR}. 

\section{Conclusions and Future Work}

To summarize, off-policy RL algorithms appear to be more robust against forgetting compared to on-policy algorithms due to their use of replay. The use of model-based RL on the other hand, does not appear to provide any direct benefit to preventing forgetting. As a result, off-policy RL algorithms are likely better suited for cyber-physical systems and other real-world applications, where the environment is non-stationary. 
However, the use of replay did not fully prevent forgetting in the SAC agent, it only delayed and slowed down the problem. Overall, all three of the evaluated algorithms suffered from catatastrophic forgetting, harmful biases from previous tasks, as well as dependency on the reward model and hyperparameters. 

Based on this, we identify the following directions for future work in CRL: Firstly, the sensitivity to hyperparameters means that it is important to investigate how to select hyperparameters that are robust to all future (unknown) tasks. Alternative approaches might be defining algorithms that do not rely on hyperparameters or enabling CRL agents to perform autotuning at runtime.
Similarly, future work should also consider how the environment can be abstracted so that these abstractions are suitable for all future (unknown) tasks. This may include enabling agents to identify and adapt to new state or action spaces. Generalizable reward functions and intrinsic rewards \cite{Gottlieb_2013_Curiosity} may also be interesting avenues to address the robustness of reward models.
Finally, more emphasis needs to be placed on resource efficiency. Not only is there a trade-off between stability and plasticity in existing CRL, but there also appears to be a trade-off between retaining past tasks and memory usage. However, cyber-physical systems typically have constrained resources and so it is important to find scalable and resource efficient solutions when developing CRL for real-world applications.

\begin{credits}
\subsubsection{\ackname} This publication has emanated from research conducted with the financial support of Taighde Éireann – Research Ireland under Grant number 21/FFP-A/8957.

\subsubsection{\discintname}
The authors have no competing interests to declare that are relevant to the content of this article.
\end{credits}

%
%
%
\bibliographystyle{unsrt}
\bibliography{references}

\end{document}